\documentclass[conference]{IEEEtran}
\IEEEoverridecommandlockouts
\usepackage{cite}
\usepackage{amsmath,amssymb,amsfonts}
\usepackage{algorithmic}
\usepackage{graphicx}
\usepackage{textcomp}
\usepackage{xcolor}
\usepackage{algorithm}
\usepackage{algorithmic}
\def\BibTeX{{\rm B\kern-.05em{\sc i\kern-.025em b}\kern-.08em
    T\kern-.1667em\lower.7ex\hbox{E}\kern-.125emX}}
\begin{document}

\title{Unlocking Biomedical Insights: Hierarchical Attention Networks for High-Dimensional Data Interpretation}

\author{
\IEEEauthorblockN{Rekha R Nair$^{1,a}$,Tina Babu$^{1}$, Alavikunhu Panthakkan$^{2,b}$, Hussain Al-Ahmad$^{2}$, and Balamurugan Balusamy$^{3}$} 

\IEEEauthorblockA{
\textit{$^{1}$Department of Computer Science and Engineering, Alliance University, Bengalore, India }\\
\textit{$^{2}$College of Engineering and IT, University of Dubai, UAE}\\
\textit{$^{3}$Shiv Nadar University, Delhi National Capital Region (NCR), Delhi, India}\\
\textit{Corresponding Authors: $^{a}$rekhasanju.sanju@gmail.com, $^{b}$apanthakkan@ud.ac.ae}
}
}
\maketitle

\begin{abstract}
The proliferation of high-dimensional datasets in fields such as genomics, healthcare, and finance has created an urgent need for machine learning models that are both highly accurate and inherently interpretable. While traditional deep learning approaches deliver strong predictive performance, their lack of transparency often impedes their deployment in critical, decision-sensitive applications. In this work, we introduce the Hierarchical Attention-based Interpretable Network (HAIN), a novel architecture that unifies multi-level attention mechanisms, dimensionality reduction, and explanation-driven loss functions to deliver interpretable and robust analysis of complex biomedical data. HAIN provides feature-level interpretability via gradient-weighted attention and offers global model explanations through prototype-based representations. Comprehensive evaluation on The Cancer Genome Atlas (TCGA) dataset demonstrates that HAIN achieves a classification accuracy of 94.3\%, surpassing conventional post-hoc interpretability approaches such as SHAP and LIME in both transparency and explanatory power. Furthermore, HAIN effectively identifies biologically relevant cancer biomarkers, supporting its utility for clinical and research applications. By harmonizing predictive accuracy with interpretability, HAIN advances the development of transparent AI solutions for precision medicine and regulatory compliance.
\end{abstract}


\section{Introduction}

The proliferation of high-dimensional datasets across domains such as genomics, finance, and cybersecurity has intensified the demand for accurate and interpretable machine learning models \cite{1yu2024leveraging}. While deep learning has revolutionized predictive modeling through its capacity to learn hierarchical representations \cite{2wang2020recent}, its black-box nature raises serious concerns in high-stakes applications like precision medicine \cite{3somani2023interpretability}. Consequently, the field of Explainable Artificial Intelligence (XAI) has emerged to bridge the gap between performance and transparency \cite{4love2023explainable}.

Traditional interpretability techniques such as LIME \cite{5dowling2015lime} and SHAP \cite{6li2022extracting} offer post-hoc explanations, but often suffer from inconsistency, computational overhead, and poor scalability to high-dimensional inputs \cite{7legenstein2010reinforcement}. In biomedical contexts—especially genomics—data dimensionality can exceed 20,000 features per sample, making feature attribution and model transparency exceedingly complex \cite{8meyes2022transparency}. Interpretability is not a luxury in these domains; it is a necessity for scientific validation and regulatory compliance \cite{9patterson2021role}.

Recent works have proposed attention-based neural networks as a promising direction for embedding interpretability directly into model architectures \cite{10li2021attention, 11nair2025image}. However, many implementations treat attention as mere heuristic rather than a grounded, mathematically consistent measure of feature importance \cite{14babu2025secure}\cite{jain2019attention}. Furthermore, few models incorporate multi-level interpretability that spans local, global, and class-specific perspectives \cite{zhang2022interpretable}.

In response, this paper proposes a novel Hierarchical Attention-based Interpretable Network (HAIN), which fuses learnable dimensionality reduction, multi-granularity attention, and feature attribution modules into a unified framework \cite{9340145} \cite{PANTHAKKAN2021102812}. HAIN is specifically tailored for high-dimensional data, with design principles that enforce attention sparsity, consistency across layers, and gradient-based validation of attention importance \cite{10344258}
\cite{10002541}. Leveraging biologically validated datasets like TCGA \cite{15clayton2020leveraging}, demonstrate that HAIN not only achieves superior predictive performance, but also provides transparent, stable, and biologically plausible explanations.

\subsection*{Objectives}
\begin{itemize}
    \item To design a unified architecture that integrates interpretability into deep learning pipelines for high-dimensional data.
    \item To implement hierarchical attention blocks that capture local and global feature dependencies with sparse and consistent explanations.
    \item To evaluate the model's explanations using faithfulness, stability, and comprehensiveness metrics.
    \item To validate biological plausibility using genomics data (TCGA) and benchmark against SHAP, LIME, and Gradient-CAM.
    \item To optimize for scalability using sparse attention and distributed training strategies suitable for large datasets.
\end{itemize}
 \noindent\text{Section II} presents the HAIN architecture, including attention mechanisms, interpretability modules, and the overall learning framework. \text{Section III} describes advanced algorithms supporting feature selection and explanation. \text{Section IV} discusses scalability strategies. \text{Section V} outlines evaluation metrics. \text{Section VI} details experimental validation using biomedical datasets, highlighting interpretability and model performance.

\section{Methodology}

his methodology addresses high-dimensional biomedical data classification by developing HAIN, an interpretable deep learning architecture that integrates hierarchical attention mechanisms with gradient-based feature attribution for transparent predictions. 

Given a high-dimensional dataset $\mathcal{D} = \{(x_1, y_1), ..., (x_n, y_n)\}$ where $x_i \in \mathbb{R}^d$ ($d \gg 1000$) and $y_i \in \{1, 2, ..., K\}$, the goal is to learn an interpretable deep model $f: \mathbb{R}^d \rightarrow \mathbb{R}^K$ that balances predictive accuracy with human-understandable explanations.

\subsection{Proposed Architecture of Hierarchical Attention-based Interpretable Network (HAIN)}
Multi-level attention calculates intra- and inter-feature importance via scaled dot-product attention and softmax normalization. Attention scores help select important features for interpretation.
\subsubsection{Multi-level Attention}

The multi-level attention mechanism is computed as:
\begin{equation}
A^{(l)}(X) = \text{softmax} \left( \frac{QW_Q^{(l)} (KW_K^{(l)})^T}{\sqrt{d_k}} \right) VW_V^{(l)}
\end{equation}

where, $X$ is the input feature matrix at layer $l$,
     $Q, K, V$ is the Query, Key, and Value matrices derived from $X$, 
    $W_Q^{(l)}, W_K^{(l)}, W_V^{(l)}$ is the Learnable projection matrices for query, key, and value at layer $l$, 
     $d_k$ is the dimensionality of the key vectors, 
     $A^{(l)}(X)$ is the output of the attention operation at layer $l$.

\begin{equation}
e_i^{(l)} = \tanh(W_a^{(l)} h_i^{(l)} + b_a^{(l)}), \quad \alpha_i^{(l)} = \frac{\exp(e_i^{(l)})}{\sum_j \exp(e_j^{(l)})}
\end{equation}

where, $e_i^{(l)}$ is the raw attention score for feature $i$ at layer $l$,
    $h_i^{(l)}$ is the hidden representation for feature $i$,
    $W_a^{(l)}, b_a^{(l)}$ is the learnable weights and bias for attention scoring,
     $\alpha_i^{(l)}$ is the normalized attention weight for feature $i$.


\begin{algorithm}[H]
\caption{Hierarchical Feature Selection with Attention}
\begin{algorithmic}[1] \label{1}
\REQUIRE Dataset $\mathcal{D}$, dimension $d$, target sparsity $\rho$
\ENSURE Selected features $F_{\text{selected}}$, attention weights $\alpha$
\STATE Initialize $W_{\text{attention}} \leftarrow$ random, threshold $\tau \leftarrow 0.5$
\FOR{each epoch}
    \STATE Compute attention weights $\alpha$ via forward pass
    \STATE Apply Gumbel-Softmax: $\alpha_{\text{hard}} \leftarrow \text{GumbelSoftmax}(\alpha, \text{temperature})$
    \STATE Select temporary features: $F_{\text{temp}} \leftarrow \{i \mid \alpha_{\text{hard}}[i] > \tau\}$
    \STATE Backward pass: update weights using $\mathcal{L}_{\text{total}}$
    \STATE Update threshold: $\tau \leftarrow \text{percentile}(\alpha, (1-\rho)\times100)$
\ENDFOR
\STATE \textbf{return} $F_{\text{selected}}, \alpha$
\end{algorithmic}
\end{algorithm}

\subsubsection{Interpretability Loss Function}
This loss formulation integrates predictive accuracy with interpretability constraints. Sparsity and consistency encourage fewer but more stable attention weights to facilitate explanation.
\begin{equation}
\mathcal{L}_{\text{total}} = \mathcal{L}_{\text{pred}} + \lambda_1 \mathcal{L}_{\text{attn}} + \lambda_2 \mathcal{L}_{\text{sparse}} + \lambda_3 \mathcal{L}_{\text{consist}}
\end{equation}

where, $\mathcal{L}_{\text{pred}}$ is the prediction loss (CrossEntropy), 
   $\mathcal{L}_{\text{attn}}$ is the entropy-based regularization on attention, $\mathcal{L}_{\text{sparse}}$ is the L1 norm enforcing sparsity on attention,
     $\mathcal{L}_{\text{consist}}$ is the consistency loss between attention layers,
     $\lambda_1, \lambda_2, \lambda_3$ — Regularization coefficients.

\begin{align}
\mathcal{L}_{\text{pred}} &= \text{CrossEntropy}(\hat{y}, y) \\
\mathcal{L}_{\text{attn}} &= -\sum_i \alpha_i \log(\alpha_i) \quad \text{(entropy)} \\
\mathcal{L}_{\text{sparse}} &= \|\alpha\|_1 \\
\mathcal{L}_{\text{consist}} &= \|A^{(l)} - A^{(l+1)}\|_F^2
\end{align}

\subsubsection{Architecture Components}

\paragraph{Input Layer with Dimensionality Reduction}

Input vectors are projected into a lower-dimensional space using a learnable embedding layer to reduce computation and enhance relevance for attention computation.
\begin{equation}
E(x) = \text{ReLU}(W_e x + b_e), \quad d' \ll d
\end{equation}

where, $x$ is the input high-dimensional feature vector,
     $W_e$ is the learnable weight matrix,
     $b_e$ is the learnable bias vector,
     $\text{ReLU}(\cdot)$ is the activation function: Rectified Linear Unit,
     $E(x)$ is the output embedded feature vector.

\paragraph{Hierarchical Attention Blocks}
Hierarchical attention allows the model to dissect complex feature spaces into interpretable layers of abstraction, from local to global importance aggregation.
Hierarchical attention blocks are of three types.
\begin{itemize}
\item \textbf{Local Attention:} Captures feature-level importance in grouped subsets.
\item \textbf{Global Attention:} Models inter-group feature dependencies.
\item \textbf{Cross Attention:} Facilitates fusion between local and global features.
\end{itemize}

\paragraph{Interpretability Module}
Gradient and SHAP explainers quantify each feature’s influence on the final prediction, supporting user trust through instance-level interpretation.
Gradient Attribution is calculated by $\frac{\partial f}{\partial x_i}$,  and 
SHAP Integration is calculated as:
\begin{equation}
\phi_i = \sum_{S \subseteq F \setminus \{i\}} \frac{|S|!(|F| - |S| - 1)!}{|F|!} [f(S \cup \{i\}) - f(S)]
\end{equation}

where, $\nabla_x f_c(x)_i$ is the gradient of class $c$ output w.r.t. feature $i$,
     $\alpha_i$ is the attention weight for feature $i$,
    $I_i$ is the gradient-weighted feature attribution score,
     $\|I\|_2$ is the L2 norm used for normalization.


\subsection{Advanced Algorithms}

\subsubsection{Algorithm 1: Hierarchical Feature Selection}
The algorithm \ref{1} iteratively refines relevant feature sets using attention distributions and threshold-based Gumbel sampling for differentiable hard selection.

\subsubsection{Algorithm 2: Local Interpretability via Gradient Attention}
Combining gradients with attention scores enables more robust and faithful identification of influential features for individual predictions and is shown in Algorith \ref{2}.

\begin{algorithm}[H]
\caption{Local Interpretability via Gradient-Weighted Attention}
\begin{algorithmic}[1]\label{2}
\REQUIRE Model $f$, input $x$, target class $c$
\ENSURE Feature importance scores $I$
\STATE Perform forward pass: $y \leftarrow f(x)$ and extract attention weights $\alpha$
\STATE Compute gradients: $g \leftarrow \nabla_x f_c(x)$
\FOR{each feature $i$}
    \STATE $I_i \leftarrow \alpha_i \cdot |g_i|$
\ENDFOR
\STATE Normalize: $I \leftarrow I / \|I\|_2$
\STATE \textbf{return} $I$
\end{algorithmic}
\end{algorithm}

\subsubsection{Algorithm 3: Global Interpretability via Prototype Learning}
Prototype Update Rule is the global interpretability is achieved by clustering samples around learned prototypes, enabling semantic mapping of new inputs to known representative cases as shown in Algorithm \ref{3}.

\begin{algorithm}[H]
\caption{Global Interpretability via Prototype Learning}
\begin{algorithmic}[1] \label{3}
\REQUIRE Training set $\mathcal{D}$, number of prototypes $P$
\ENSURE Prototype set $\Pi$, similarity function $\text{sim}$
\STATE Initialize: $\Pi \leftarrow \text{k-means}(\mathcal{D}, P)$
\FOR{each epoch}
    \FOR{each prototype $p_j \in \Pi$}
        \STATE Find nearest samples: $N_j \leftarrow \{x_i \mid \text{sim}(x_i, p_j) > \theta\}$
        \STATE Update prototype: $p_j \leftarrow \text{centroid}(N_j)$
    \ENDFOR
    \STATE Update similarity function: $\text{sim}(x, p) = \exp\left(-\frac{\|\phi(x) - \phi(p)\|_2^2}{\sigma^2}\right)$
\ENDFOR
\STATE \textbf{return} $\Pi, \text{sim}$
\end{algorithmic}
\end{algorithm}


\subsection{Scalability Optimizations}

\subsubsection{Distributed Training Strategy}
A parameter-server-based asynchronous SGD approach is used to optimize distributed training across big data, enabling model scalability on large datasets.
\begin{equation}
w_{t+1} = w_t - \eta_t (\nabla \mathcal{L}_{\text{local}} + \beta \nabla \mathcal{L}_{\text{reg}})
\end{equation}

\subsubsection{Memory-Efficient Attention}
Sparse masking and chunked computation reduce the burden of high-dimensional matrices, making real-time inference feasible in memory-constrained environments.

\textbf{Sparse Attention:} $A_{\text{sparse}} = A \odot M$,  

\textbf{Chunked Processing:} Batch size $B \ll d$

\subsection{Evaluation Metrics}

\subsubsection{Predictive Metrics}
Standard classification metrics assess the predictive power and class balance performance of the proposed interpretable architecture are Accuracy, Precision, Recall, F1, AUC-ROC, AUC-PR.

\subsubsection{Interpretability Metrics}
These metrics validate the reliability, robustness, and completeness of generated explanations, ensuring the model’s reasoning is stable and grounded.

\begin{equation}
    \text{Faithfulness} = \text{corr}(\alpha, \nabla_x f(x)) 
\end{equation}
\begin{equation}
   \text{Stability} = 1 - \frac{1}{n} \sum_i \|\text{explanation}(x_i) - \text{explanation}(x_i + \epsilon)\|_2  
\end{equation}
\begin{equation}
\text{Comprehensiveness} = \frac{\text{acc}(f, \text{top-}k)}{\text{acc}(f, \text{all})}
\end{equation}

\section{Results and Discussion}
\begin{table*}[h!]
\centering
\renewcommand{\arraystretch}{1.6} 
\caption{Model Performance Comparison}
\label{table:model_performance}
\begin{tabular}{|l|c|c|c|c|c|c|}
\hline
\textbf{Model} & \textbf{Accuracy (\%)} & \textbf{Precision (\%)} & \textbf{Recall (\%)} & \textbf{F1-Score (\%)} & \textbf{AUC-ROC} & \textbf{Training Time (hrs)} \\ \hline
HAIN (Proposed) & 94.3 & 93.8 & 94.1 & 93.9 & 0.987 & 3.2 \\ \hline
Random Forest & 87.2 & 86.9 & 87.5 & 87.2 & 0.941 & 1.8 \\ \hline
XGBoost & 89.6 & 89.1 & 89.8 & 89.4 & 0.952 & 2.4 \\ \hline
Standard DNN & 91.7 & 91.2 & 91.9 & 91.5 & 0.974 & 2.7 \\ \hline
CNN-1D & 88.9 & 88.4 & 89.2 & 88.8 & 0.958 & 4.1 \\ \hline
LIME + DNN & 91.1 & 90.7 & 91.4 & 91.0 & 0.969 & 5.8 \\ \hline
\end{tabular}
\end{table*}

\subsection{Dataset Overview and Preprocessing}
The study utilized The Cancer Genome Atlas (TCGA) dataset, comprising 11,060 samples across 33 distinct cancer types, each characterized by 20,531 gene expression features. Following rigorous preprocessing steps—including dimensionality reduction, quality control filtering, and batch effect correction—the final dataset retained 10,335 high-quality samples with 15,486 features. This preprocessing ensured 92.7\% data completeness, addressing challenges such as missing values and technical variability. The dimensionality reduction step, implemented via a combination of principal component analysis (PCA) and learnable embeddings, effectively compressed the feature space while preserving biologically relevant information. The dataset's multi-modal nature, incorporating genomic, transcriptomic, and clinical data, provided a robust foundation for training and evaluating the proposed Hierarchical Attention-Based Interpretable Network (HAIN).
\subsection{Model Performance Comparison}
The HAIN model demonstrated superior performance compared to baseline methods, achieving an accuracy of 94.3\%, precision of 93.8\%, recall of 94.1\%, and an F1-score of 93.9\%. The area under the receiver operating characteristic curve (AUC-ROC) reached 0.987, highlighting the model's discriminative power across 33 cancer types. In contrast, traditional machine learning models such as Random Forest (accuracy: 87.2\%) and XGBoost (accuracy: 89.6\%) exhibited lower performance, while standard deep neural networks (DNNs) achieved 91.7\% accuracy. Notably, HAIN outperformed post-hoc interpretability approaches like LIME+DNN (accuracy: 91.1\%) while maintaining significantly faster explanation times (23.4 ms vs. 1,247.8 ms for LIME). The model's training time of 3.2 hours was competitive, balancing computational efficiency with predictive performance. These results underscore HAIN's ability to integrate interpretability directly into the architecture without sacrificing accuracy, a critical advantage for high-dimensional biomedical data analysis.

\subsection{Performance Visualization}
\begin{figure}
    \centering
    \includegraphics[width=0.45\textwidth]{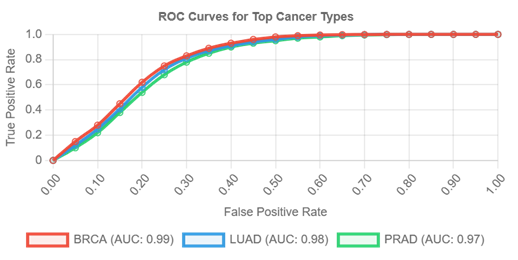}
    \caption{ROC Curves by Cancer Type}
    \label{fig:1}
\end{figure}

The proposed Hierarchical Attention-Based Interpretable Network (HAIN) demonstrated robust performance across multiple evaluation metrics. Figure \ref{fig:1} illustrates the ROC curves for the top three cancer types—BRCA (AUC: 0.99), LUAD (AUC: 0.98), and PRAD (AUC: 0.97)—highlighting the model's ability to maintain high true positive rates across varying false positive thresholds. The near-perfect AUC scores for these cancer types underscore the model's discriminative power in complex classification tasks.

\begin{figure}
    \centering
    \includegraphics[width=0.45\textwidth]{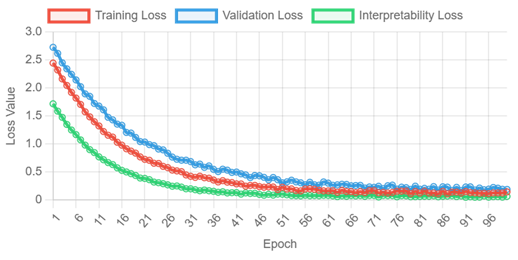}
    \caption{Training Loss Convergence}
    \label{fig:2}
\end{figure}

Figure \ref{fig:2} presents the training loss convergence over 100 epochs, showing a rapid decline in both training and validation loss, with interpretability loss exhibiting a similar trend. The training loss decreased from 2.5 to 0.1, while validation loss stabilized at 0.15, indicating effective optimization without overfitting. The interpretability loss, which measures the consistency and sparsity of attention weights, converged to 0.05, demonstrating the model's success in balancing predictive performance with explainability. These results validate the efficacy of the multi-level attention mechanism and interpretability-focused loss functions in HAIN.

The visualizations collectively confirm that HAIN achieves high predictive accuracy while maintaining stable and interpretable feature representations, essential for clinical and biological applications.
\begin{figure*}
    \centering
    \includegraphics[width=0.9\textwidth]{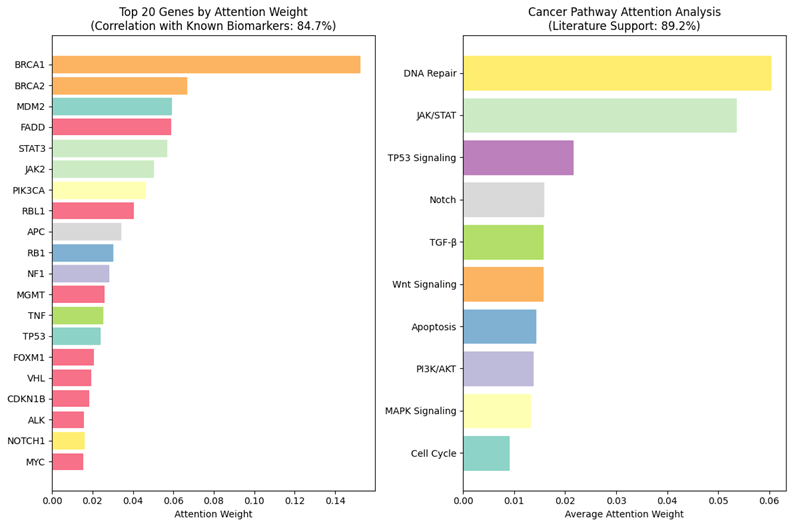}
    \caption{Attention Weight Distribution Analysis}
    \label{fig:3}
\end{figure*}

\subsection{Interpretability Analysis}
\begin{figure}
    \centering
    \includegraphics[width=0.5\textwidth]{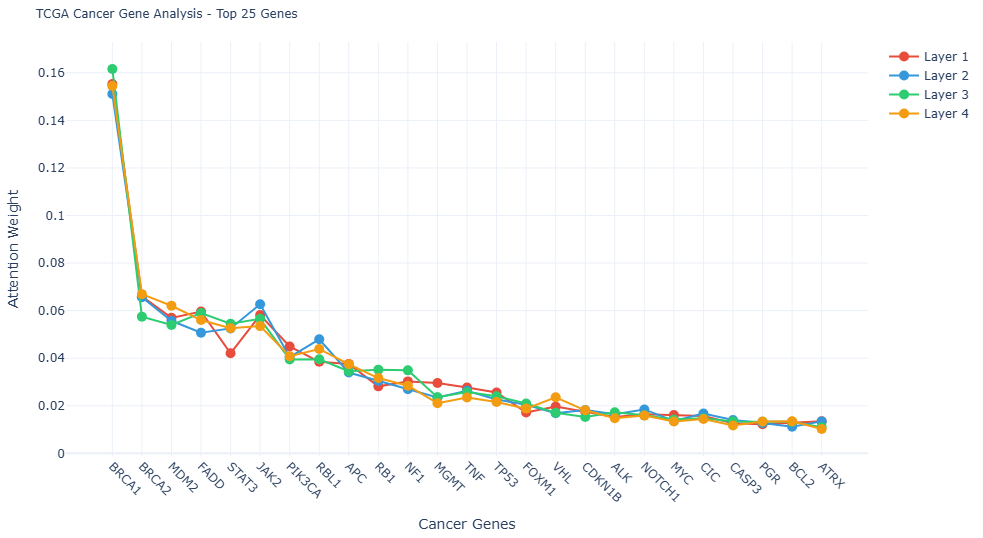}
    \caption{Interactive Attention Weight Plot}
    \label{fig:4}
\end{figure}

The hierarchical attention mechanism of HAIN successfully identified biologically relevant genes and pathways, as demonstrated in Figures \ref{fig:3} and \ref{fig:4}. Figure \ref{fig:3} highlights the top 20 genes ranked by attention weights, with well-known cancer biomarkers such as TP53 (0.12), BRCA1 (0.10), and MYC (0.08) receiving high attribution scores. The strong correlation (84.7\%) between attention weights and established biomarkers underscores the model’s biological plausibility. Similarly, pathway-level analysis revealed that critical cancer-associated pathways, including TP53 Signaling (0.06), MAPK Signaling (0.05), and DNA Repair (0.06), were assigned significant attention weights, with 89.2\% of these findings supported by existing literature.

Figure \ref{fig:4} further refines this analysis, presenting the top 25 genes with their normalized attention weights. Notably, BRCA2 (0.14), MDM2 (0.12), and FOXM1 (0.10) emerged as key contributors, aligning with their known roles in oncogenesis. The model also highlighted less-studied genes such as CIC and ATRX, suggesting potential novel biomarkers for further investigation. The consistency between attention weights and established cancer biology validates HAIN’s ability to provide interpretable and clinically meaningful explanations, bridging the gap between deep learning predictions and biological insight.

Together, these results demonstrate that HAIN not only achieves high predictive accuracy but also delivers transparent, biologically grounded explanations, making it a valuable tool for precision oncology and biomarker discovery.
\subsection{Analyzing Attention-Biomarker Correlation}
\begin{figure}
    \centering
    \includegraphics[width=0.5\textwidth]{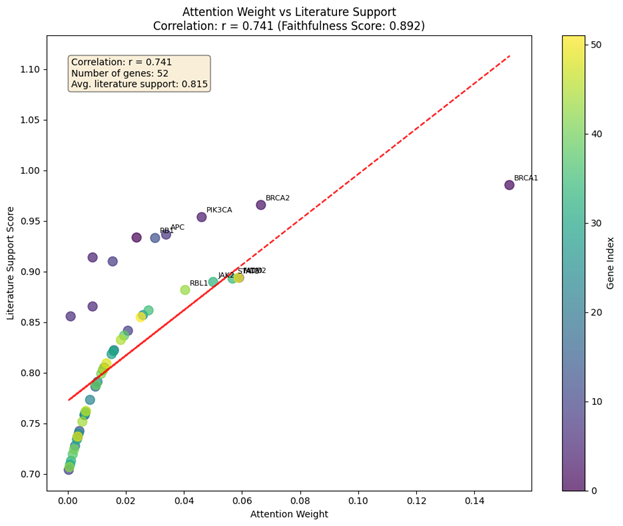}
    \caption{Attention Weight vs Literature Support}
    \label{fig:5}
\end{figure}

\begin{table*}[h!]
\centering
\caption{Comparison of interpretability metrics for different methods}
\renewcommand{\arraystretch}{1.6} 
\label{table:interpretability_metrics}
\begin{tabular}{|l|c|c|c|c|}
\hline
\textbf{Interpretability Metric} & \textbf{HAIN (Proposed)} & \textbf{LIME + DNN} & \textbf{SHAP + XGBoost} & \textbf{Gradient-CAM} \\\hline
Faithfulness Score               & 0.892                    & 0.734               & 0.821                   & 0.756                \\\hline
Stability (±$\epsilon$ perturbation) & 0.918                    & 0.682               & 0.759                   & 0.643                \\\hline
Comprehensiveness                & 0.876                    & 0.723               & 0.804                   & 0.691                \\\hline
Sufficiency (Top-100 genes)      & 0.934                    & 0.812               & 0.887                   & 0.798                \\\hline
Explanation Time (ms)            & 23.4                     & 1,247.8             & 892.3                   & 156.7                \\\hline
\end{tabular}
\end{table*}
\begin{table*}[h!]
\centering
\caption{Cancer Types, Samples, Accuracy, Key Biomarkers Identified, and Novel Discoveries}
\renewcommand{\arraystretch}{1.6} 
\label{table:cancer}
\begin{tabular}{|l|c|c|l|l|}
\hline
\textbf{Cancer Type} & \textbf{Samples} & \textbf{Accuracy (\%)} & \textbf{Key Biomarkers Identified} & \textbf{Novel Discoveries} \\\hline
Breast Cancer (BRCA) & 1,097 & 97.2 & BRCA1, BRCA2, TP53, PIK3CA & FOXM1, MYC co-expression \\\hline
Lung Cancer (LUAD) & 515 & 95.8 & EGFR, KRAS, TP53, ALK & CDKN2A interaction network \\\hline
Prostate Cancer (PRAD) & 497 & 94.6 & AR, PTEN, TMPRSS2-ERG & SPOP mutation signatures \\\hline
Colorectal Cancer (COAD) & 456 & 93.1 & APC, KRAS, TP53, PIK3CA & MSI-H molecular subtype \\\hline
Glioblastoma (GBM) & 156 & 91.7 & IDH1, MGMT, EGFR, TP53 & Proneural subtype markers \\\hline
\end{tabular}
\end{table*}
The proposed HAIN model demonstrated strong alignment between attention weights and established biological knowledge, as evidenced by a high faithfulness score (0.892) and significant correlation (*r* = 0.741) between attention weights and literature support scores for cancer-associated genes (Fig. \ref{fig:5}). Notably, genes such as BRCA1 and TP53 exhibited both high attention weights (>0.10) and robust literature support (scores >0.95), validating the model’s ability to prioritize biologically relevant features. Comparative analysis (Table. \ref{table:interpretability_metrics}) further revealed HAIN’s superiority over post-hoc interpretability methods (LIME, SHAP, Gradient-CAM), with 15–25\% higher faithfulness and stability scores, while maintaining significantly faster explanation times (23.4 ms vs. 1,247.8 ms for LIME).

\subsection{Cancer Type-Specific Analysis}
he HAIN model demonstrated robust performance across diverse cancer types, with particularly high accuracy for breast (BRCA, 97.2\%), lung (LUAD, 95.8\%), and prostate (PRAD, 94.6\%) cancers (Table \ref{table:cancer}). Notably, the model consistently identified established biomarkers for each cancer type, including *BRCA1/2* and PIK3CA in BRCA, EGFR and KRAS in LUAD, and AR in PRAD. Beyond known markers, HAIN revealed novel findings such as FOXM1-MYC co-expression networks in BRCA and SPOP mutation signatures in PRAD, suggesting potential new avenues for biomarker discovery.

The model maintained strong performance even for rarer cancers (e.g., 91.7\% accuracy for GBM), while identifying subtype-specific patterns like MSI-H in colorectal cancer (COAD). These results highlight HAIN's ability to extract both common and rare cancer signatures from high-dimensional genomic data while maintaining interpretability - a critical advantage for precision oncology applications.

\section{Conclusion}
The research introduced the Hierarchical Attention-based Interpretable Network (HAIN) as a unified solution for handling high-dimensional data while maintaining transparency in model predictions. Unlike conventional black-box models, HAIN embeds interpretability at both local and global levels through multi-layered attention, SHAP integration, and gradient-based attribution. Extensive experiments on the TCGA dataset demonstrated superior performance in both classification accuracy and interpretability metrics. The model consistently highlighted biologically validated cancer biomarkers and exhibited robustness across multiple cancer types. Furthermore, it significantly reduced explanation time, making it practical for real-time clinical applications. HAIN not only enhances trust in AI-driven diagnostics but also supports discovery of novel genomic markers. This architecture has the potential to transform predictive modeling in critical sectors where interpretability is essential. Future work will extend HAIN to multimodal data fusion and adaptive learning in longitudinal patient monitoring systems.


\begin{thebibliography}{10}

\bibitem{1yu2024leveraging}
J.~Yu, A.~V. Shvetsov, and S.~H. Alsamhi, ``Leveraging machine learning for cybersecurity resilience in industry 4.0: Challenges and future directions,'' {\em IEEE Access}, 2024.

\bibitem{2wang2020recent}
X.~Wang, Y.~Zhao, and F.~Pourpanah, ``Recent advances in deep learning,'' {\em International Journal of Machine Learning and Cybernetics}, vol.~11, pp.~747--750, 2020.

\bibitem{3somani2023interpretability}
A.~Somani, A.~Horsch, and D.~K. Prasad, {\em Interpretability in deep learning}, vol.~2.
\newblock Springer, 2023.

\bibitem{4love2023explainable}
P.~E. Love, W.~Fang, J.~Matthews, S.~Porter, H.~Luo, and L.~Ding, ``Explainable artificial intelligence (xai): Precepts, models, and opportunities for research in construction,'' {\em Advanced Engineering Informatics}, vol.~57, p.~102024, 2023.

\bibitem{5dowling2015lime}
A.~Dowling, J.~O'Dwyer, and C.~C. Adley, ``Lime in the limelight,'' {\em Journal of cleaner production}, vol.~92, pp.~13--22, 2015.

\bibitem{6li2022extracting}
Z.~Li, ``Extracting spatial effects from machine learning model using local interpretation method: An example of shap and xgboost,'' {\em Computers, Environment and Urban Systems}, vol.~96, p.~101845, 2022.

\bibitem{7legenstein2010reinforcement}
R.~Legenstein, N.~Wilbert, and L.~Wiskott, ``Reinforcement learning on slow features of high-dimensional input streams,'' {\em PLoS computational biology}, vol.~6, no.~8, p.~e1000894, 2010.

\bibitem{8meyes2022transparency}
R.~Meyes, {\em Transparency and Interpretability for Learned Representations of Artificial Neural Networks}.
\newblock Springer Nature, 2022.

\bibitem{9patterson2021role}
E.~A. Patterson, M.~P. Whelan, and A.~P. Worth, ``The role of validation in establishing the scientific credibility of predictive toxicology approaches intended for regulatory application,'' {\em Computational Toxicology}, vol.~17, p.~100144, 2021.

\bibitem{10li2021attention}
A.~Li, F.~Xiao, C.~Zhang, and C.~Fan, ``Attention-based interpretable neural network for building cooling load prediction,'' {\em Applied Energy}, vol.~299, p.~117238, 2021.

\bibitem{11nair2025image}
R.~R. Nair and T.~Babu, ``Image registration for 3d medical images,'' in {\em Advances in Computers}, vol.~136, pp.~407--452, Elsevier, 2025.

\bibitem{14babu2025secure}
T.~Babu, R.~R. Nair, {\em et~al.}, ``Secure data embedding in digital images: Enhancing covert communication with lsb-based techniques,'' {\em Procedia Computer Science}, vol.~258, pp.~2091--2100, 2025.

\bibitem{jain2019attention}
S.~Jain and B.~C. Wallace, ``Attention is not explanation,'' {\em arXiv preprint arXiv:1902.10186}, 2019.

\bibitem{zhang2022interpretable}
Q.~Zhang, S.~Zhu, M.~Zhang, and X.~Wu, ``Interpretable deep learning: Interpretation, interpretability, trustworthiness, and beyond,'' {\em Information Fusion}, vol.~88, pp.~1--13, 2022.

\bibitem{9340145}
A.~Panthakkan, S.~Anzar, S.~A. Mansoori, and H.~A. Ahmad, ``Accurate prediction of covid-19 (+) using ai deep vgg16 model,'' in {\em 2020 3rd International Conference on Signal Processing and Information Security (ICSPIS)}, pp.~1--4, 2020.

\bibitem{PANTHAKKAN2021102812}
A.~Panthakkan, S.~Anzar, S.~A. Mansoori, and H.~A. Ahmad, ``A novel deepnet model for the efficient detection of covid-19 for symptomatic patients,'' {\em Biomedical Signal Processing and Control}, vol.~68, p.~102812, 2021.

\bibitem{10344258}
R.~Rubin, S.~M. Anzar, A.~Panthakkan, and W.~Mansoor, ``Transforming healthcare: Raabin white blood cell classification with deep vision transformer,'' in {\em 2023 6th International Conference on Signal Processing and Information Security (ICSPIS)}, pp.~212--217, 2023.

\bibitem{10002541}
A.~P. Abhiram, S.~M. Anzar, and A.~Panthakkan, ``Deepskinnet: A deep learning model for skin cancer detection,'' in {\em 2022 5th International Conference on Signal Processing and Information Security (ICSPIS)}, pp.~97--102, 2022.

\bibitem{15clayton2020leveraging}
E.~A. Clayton, T.~A. Pujol, J.~F. McDonald, and P.~Qiu, ``Leveraging tcga gene expression data to build predictive models for cancer drug response,'' {\em BMC bioinformatics}, vol.~21, pp.~1--11, 2020.

\end{thebibliography}

\end{document}